\documentclass{article}

\usepackage{arxiv}

\usepackage[utf8]{inputenc} 
\usepackage[T1]{fontenc}    
\usepackage{hyperref}       
\usepackage{url}            
\usepackage{booktabs}       
\usepackage{amsfonts}       
\usepackage{nicefrac}       
\usepackage{microtype}      
\usepackage{lipsum}		
\usepackage{graphicx}
\usepackage{doi}

\usepackage[all]{nowidow}
\usepackage{doi}
\usepackage{xcolor}
\usepackage{tikz}
\usetikzlibrary{shapes, arrows, positioning, backgrounds, fit, calc}
\usepackage{multirow}
\usepackage{tabularx,siunitx}
\usepackage{float}
\usepackage{booktabs}
\usepackage[numbers]{natbib}

\title{NormEnsembleXAI: Unveiling the~Strengths and~Weaknesses of XAI Ensemble Techniques}

\date{} 					

\author{ \href{https://orcid.org/0000-0003-2903-6050}{\includegraphics[scale=0.06]{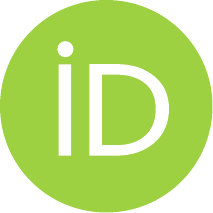}\hspace{1mm}Weronika Hryniewska-Guzik}, \hspace{1mm}
Bartosz Sawicki, \hspace{1mm}
\href{https://orcid.org/0000-0001-8423-1823}{\includegraphics[scale=0.06]{orcid.pdf}\hspace{1mm}Przemysław Biecek} \\
	Faculty of Mathematics and Information Science\\
	Warsaw University of Technology\\
	Koszykowa 75, 00-662 Warsaw (Poland) \\
	\texttt{weronika.hryniewska.dokt@pw.edu.pl} \\
}

\date{}

\renewcommand{\headeright}{Weronika Hryniewska-Guzik et al.}
\renewcommand{\undertitle}{}
\renewcommand{\shorttitle}{NormEnsembleXAI}

\hypersetup{
pdftitle={A template for the arxiv style},
pdfsubject={q-bio.NC, q-bio.QM},
pdfauthor={David S.~Hippocampus, Elias D.~Striatum},
pdfkeywords={First keyword, Second keyword, More},
}

\begin{document}
\maketitle

\begin{abstract}
This paper presents a comprehensive comparative analysis of explainable artificial intelligence~(XAI) ensembling methods. Our research brings three significant contributions. Firstly, we introduce a novel ensembling method, NormEnsembleXAI, that leverages minimum, maximum, and average functions in conjunction with normalization techniques to enhance interpretability. Secondly, we offer insights into the strengths and weaknesses of XAI ensemble methods. Lastly, we provide a library, facilitating the practical implementation of XAI ensembling, thus promoting the adoption of transparent and interpretable DL models.
\end{abstract}


\begin{figure}[h]
  \centering
  \includegraphics[width=0.6\linewidth]{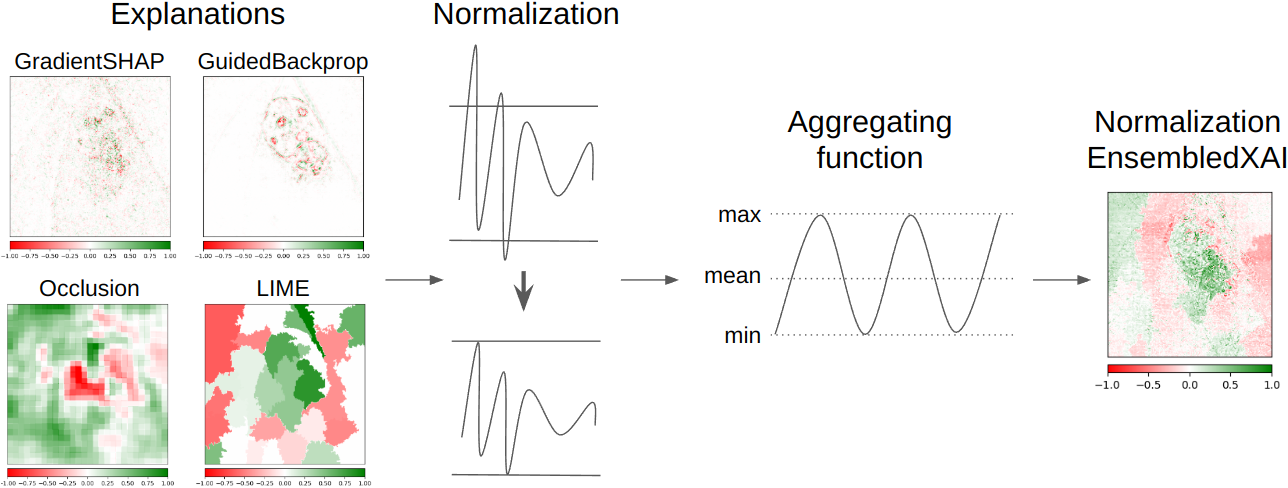}

  \caption{Visualization of the proposed NormEnsembleXAI method. The NormEnsembleXAI method is designed to handle the diverse value ranges produced by various explanation methods. To address this, the algorithm employs normalization techniques, including Second Moment Scaling, Normal Standardization, or Robust Standardization. Subsequently, it utilizes aggregation functions for ensembling explanations, including Maximum, Minimum, and Mean.}
  \label{fig:method_summary}
\end{figure}

\section{Introduction}

The growing integration of deep learning (DL) models into critical decision-making processes has underscored the need for transparency and interpretability~\cite{molnar2022, Salahuddin2022, Meyes2022}. As the use of DL models becomes more widespread, so does the concern about hidden biases and unreliable predictions~\cite{pmlr-v206-zhang23g}, making model explainability a~crucial consideration in modern machine learning applications~\cite{app11115088,Burkart2021}. Over the years, researchers have developed numerous explanation methods, each providing a~unique perspective on the underlying models' decision-making processes~\cite{BARREDOARRIETA202082, Minh2022, LOH2022107161}. However, a~single explanation often falls short of offering a~comprehensive understanding of model behavior~\cite{Kokhlikyan2020}.

As humans have limitations in processing a~substantial amount of new information simultaneously~\cite{sweller2003evolution}, there has been a~compelling argument for explanations that encompass only pertinent and representative features~\cite{DBLP:journals/corr/abs-1901-10040}. Consequently, explanations should prioritize simplicity, avoiding unnecessary complexity by utilizing a~limited set of features~\cite{10.1177/0959354316684043, Kulesza2013}. To address this limitation, ensembling methods for explanations have emerged as a~promising solution, offering a~holistic view of model behavior by combining multiple explanations~\cite{Zou2022}. 

Explanations ensembling increases explanation's robustness \cite{Rieger2020} by helping to reduce the impact of individual explanation's weaknesses or biases. By combining different explanations, the ensemble is less likely to be affected by the limitations of any single explanation. Moreover, ensembling different explanations, which might specialize in different aspects of the model's interpretability, can lead to a more comprehensive model's understanding.

However, despite the growing popularity of ensembling methods for explainability~\cite{Zou2022, Bobek2021, Rieger2020}, a~comprehensive understanding of the advantages and disadvantages of different eXplainable AI (XAI) ensembling approaches, and when to use them, remains unknown. This knowledge gap hinders the informed selection of ensembling methods for specific use cases and poses a~challenge to the broader adoption of these methods in practical applications.

Moreover, existing XAI ensembling methods have not undergone a comprehensive evaluation across multiple dimensions, such as faithfulness to the model~\cite{10.5555/3327757.3327875}, robustness to perturbations~\cite{10.5555/3454287.3455271}, localization of explanations~\cite{Zhang2018}, computational complexity~\cite{pmlr-v119-chalasani20a}, randomization properties~\cite{pmlr-v119-sixt20a}, and adherence to axiomatic principles~\cite{pmlr-v70-sundararajan17a}. These facets are crucial for assessing the reliability and suitability of ensembling techniques in diverse real-world scenarios.

In light of these challenges, we were the first to identify and fill the knowledge gap. We comprehensively analyze existing XAI ensemble methods to shed light on their relative strengths and weaknesses. By systematically evaluating these methods across a~spectrum of evaluation criteria, we aim to provide valuable insights into their practical utility and guide practitioners in choosing the most appropriate ensembling approach for their specific needs.

Furthermore, our investigation has uncovered a~niche that remains unfilled, particularly in the context of aggregation methods. Specifically, we identify an opportunity to use minimum, maximum, and average functions in conjunction with normalization techniques to enhance the interpretability of DL models (presented in Figure~\ref{fig:method_summary}). This novel approach, which has not been extensively explored in the prior literature such as~\cite{Rieger2020}, represents a~valuable contribution to the field of XAI.

In addition to our comparative analysis and novel findings, we are pleased to offer EnsembleXAI, a comprehensive PyTorch-compatible library accompanying this paper. This library is designed to facilitate the practical implementation of XAI ensembling methods in real-world machine learning applications, further advancing the adoption of interpretable DL models in critical decision-making processes. 

\section{Related works}
\label{sec:related_works}
\paragraph{Explanation methods}
Various techniques have been developed to analyze the decision-making processes of deep learning models~\cite{10.1145/3531146.3534639}. Model-agnostic methods aim to provide explanations for a~wide range of machine learning models without making specific assumptions about the model's architecture~\cite{darias2021systematic}. Post-hoc methods generate explanations after the model has been trained and do not influence the model's training process~\cite{10.1145/3546577}. They are applied to the model as an additional step to provide interpretability. Local explanation methods focus on explaining the predictions of a~model for a~specific input instance~\cite{NEURIPS2018_b495ce63}. They provide insights into why a~model made a~particular decision for a~given input without considering the model's global behavior. These techniques can be categorized into two main types based on their method of operation: gradient-based and perturbation-based methods~\cite{pmlr-v139-agarwal21c}.

Gradient-based methods, including Integrated Gradients~\cite{pmlr-v70-sundararajan17a}, Saliency~\cite{simonyan2014deep}, Gradient SHAP~\cite{NIPS2017_8a20a862}, and Guided Backpropagation~\cite{SpringenbergDBR14}, Deconvolution~\cite{zeiler2013visualizing}, Input X Gradient~\cite{10.5555/3305890.3306006}, leverage the gradients of the model's output concerning its input features to assess feature importance. These methods are beneficial for deep learning models.

Perturbation-based methods, such as LIME~\cite{Ribeiro2016}, Occlusion~\cite{zeiler2013visualizing}, and Shapley Value Sampling~\cite{CASTRO20091726}, SHAP~\cite{NIPS2017_8a20a862}, Feature Ablation~\cite{Kokhlikyan2020}, KernelSHAP~\cite{10.5555/3295222.3295230}, introduce controlled perturbations to input data and observe their effects on model predictions to determine feature importance. This perturbation is achieved through occlusion, feature substitution, masking, or conditional sampling, allowing for an analysis of how feature changes impact model predictions. 

In addition to both categories mentioned above, some techniques that do not fit neatly into one category. For example, Noise Tunnel~\cite{10.5555/3454287.3455160} is a~method that introduces controlled noise into the input data and observes the model's sensitivity to this noise, thereby providing insights into feature importance.

Two papers~\cite{Zou2022, Bobek2021} focus on the ensembling of the previously mentioned explanations. In~\citet{Zou2022}, they demonstrate their XAI ensemble algorithm in the context of localizing Covid-19 lesions. Their approach combines SHAP~\cite{CASTRO20091726} and Grad-CAM++~\cite{8354201} and evaluates its performance using metrics introduced by them. Meanwhile,~\citet{Bobek2021} combine SHAP~\cite{CASTRO20091726}, LIME~\cite{Ribeiro2016}, and Anchors~\cite{Ribeiro_Singh_Guestrin_2018} into a~unified ensemble explanation. They introduce an algorithm that assigns weights to explanations based on a~selected metric.

Our contribution builds upon both of these works, expanding their functionality with a~diverse range of explanation methods. Additionally, we incorporate essential normalization techniques and package them into a~unified library for user-friendly access and utilization.


\paragraph{XAI libraries}
In addition to research on creating new explanation methods, the field of XAI has developed many libraries and frameworks designed to facilitate the practical implementation and evaluation of explainability methods~\cite{xair, Bodria2023}. Among the XAI libraries, designed for images, are Captum for PyTorch~\cite{Kokhlikyan2020}, iNNvestigate for Keras and TensorFlow 2~\cite{Alber2019}, OmniXAI for PyTorch and Tensorflow~\cite{wenzhuo2022-omnixai}, Xplique for TensorFlow with PyTorch wrapper~\cite{fel2022xplique}, and M3d-CAM for 3D/2D PyTorch medical models~\cite{2007.00453}. Notably, it is worth mentioning that, to our knowledge, none of these libraries currently offer the capability to ensemble image explanations, a~gap that our research aims to address. 

\paragraph{Evaluation of XAI methods}
The assessment and validation of explainability methods are important in ensuring their efficacy and trustworthiness in enhancing the transparency and interpretability of machine learning models~\cite{VEERAPPA2022101539, 10.1145/3351095.3372870}. 

One of the primary methods for assessing the explanations is qualitative evaluation. It involves seeking input from human experts who review and judge the explanations for their context, coherence, and controllability~\cite{Nauta2023}. Qualitative evaluation is often conducted through user studies, surveys, or expert interviews, where the emphasis is on the interpretability of the explanations~\cite{Zhou2021}. Researchers collect valuable feedback from domain experts and end-users to understand how well the explanations align with human mental models and decision-making processes~\cite{hong2020human}.

In addition to qualitative assessments, quantitative metrics provide a more objective way to measure the performance of XAI methods~\cite{10.1145/3583558}. Common quantitative metrics include Faithfulness, Robustness, Localization, Complexity, and Randomization~\cite{Pahde_2023_CVPR}. Faithfulness measures how closely the explanation reflects the model's behavior, ensuring that the explanation accurately represents what the model has learned~\cite{10.5555/3327757.3327875}. Robustness assesses the explanation's ability to maintain its quality under various conditions, such as changes in input data or perturbations, ensuring that the explanation remains reliable~\cite{10.5555/3454287.3455271}. Localization evaluates how well the explanation pinpoints the specific features or elements in the input data that contributed to the model's decision or prediction, enhancing its interpretability~\cite{Zhang2018}. Complexity measures the simplicity and comprehensibility of the explanation, ensuring that it is not overly convoluted or difficult for users to understand~\cite{pmlr-v119-chalasani20a}. Randomization checks if the explanation remains consistent and does not exhibit significant variations when similar inputs are provided, enhancing its stability and consistency~\cite{pmlr-v119-sixt20a}.

Another important aspect of evaluation is the comparative analysis of different XAI methods. It involves putting multiple explanation methods against one another to determine which provides more informative, reliable, and understandable explanations~\cite{10.1145/3397481.3450650}. Comparative studies can help identify the strengths and weaknesses of different methods in various contexts and domains~\cite{DING2022238}.

\paragraph{XAI evaluation libraries}

To meet the demand for the~trustworthiness of explanations, a growing set of XAI evaluation tools and frameworks has emerged, offering both quantitative and comparative methods for scrutinizing the quality and performance of explainability techniques. These tools play a crucial role in assessing the transparency and interpretability of machine learning models, ultimately facilitating informed decision-making in model selection and explanation.

One noteworthy contribution is Quantus~\cite{Hedstrom2022}, an open-source toolkit designed to quantitatively evaluate explanations of neural network predictions. It addresses the need for transparency and reproducibility by providing a~comprehensive collection of evaluation metrics.

Another key framework in the field of XAI evaluation is CLEVR-XAI~\cite{Arras2022}, a~benchmark dataset for the ground truth evaluation of neural network explanations. This benchmark dataset aims to objectively measure the quality of XAI methods by focusing on the visual question-answering task. By providing a~selective, controlled, and realistic testing benchmark, CLEVR-XAI contributes to the better assessment of XAI methods.

Meanwhile, the XAI-TRIS~\cite{Clark2023} focuses on developing benchmark datasets and quantitative metrics for evaluating the performance of XAI methods, particularly in non-linear classification scenarios. This research sheds light on the challenges and differences in the explanations derived from various model architectures, offering valuable insights into XAI evaluation.

Collectively, these works contribute to the advancement of XAI evaluation and benchmarking, fulfilling the critical requirements for transparency, reproducibility, and objective assessment in the field. In our research, we introduce a novel XAI ensembling method and incorporate several metrics in our EnsembleXAI library. For NormEnsembleXAI evaluation in the image classification task, we rely on Quantus as a foundational assessment tool.



\section{NormEnsembleXAI method}\label{sec:XAI_ensemblings}

We assume the following notation, $\phi^{e\rightarrow m}_{i, j}$ is the importance of feature $j$ of the instance $i$ given by the explanation method $e$ of machine learning model $m$. By $E_i$, we denote the set of all explanations available for the instance $i$.

We compare three normalization methods. A classic approach: normal standardization by subtracting the mean and dividing by the standard deviation. The abovementioned statistics are calculated separately for each explanation algorithm $e$ across all instances $k$, and all features $l$, as presented in

\begin{equation}\label{eq:normalize_mean_variance}
\phi'^{e \rightarrow m}{i, j} = \frac{ \phi^{e \rightarrow m}{i, j} - mean_{k, l}( \phi^{e \rightarrow m}{k, l})}{std{k, l} (\phi^{e \rightarrow m}_{k, l})}.
\end{equation}

To increase robustness to outlier attributions, we tested robust standardization by subtracting the median and dividing by the interquartile range. Statistics were calculated analogously to Equation~\ref{eq:normalize_mean_variance}. The formula for this normalization method is given in

\begin{equation}\label{eq:normalize_median_iqr}
\phi'^{e \rightarrow m}{i, j} = \frac{ \phi^{e \rightarrow m}{i, j} - median_{k, l}( \phi^{e \rightarrow m}{k, l})}{IQR{k, l} (\phi^{e \rightarrow m}_{k, l})}.
\end{equation}

Another considered normalization algorithm is scaling by the average second-moment estimate~\cite{hedstrom2023metaquantus}. In this case, the averaging is done with respect to the channel dimension, and the second moment is estimated by the standard deviation. The transformation is described in

\begin{equation}\label{eq:second_moment_estimate}
\phi'^{e \rightarrow m}{i, j} = \frac{ \phi^{e \rightarrow m}{i, j}}{avg_{channels}( std_{k, l}(\phi^{e \rightarrow m}_{k, l}))}.
\end{equation}

NormEnsembleXAI method uses the aggregating function $f()$ to ensemble explanations, such as maximum, minimum, and mean. Due to the varying ranges of values across explanation methods, a~normalization $norm$ may need to be performed to obtain normalized explanations $\phi'^{e \rightarrow m}_{i, j}$. The ensemble attribution of $j$-th feature is given by

\begin{equation}\label{eq:NormEnsembleXAI_ensemble}
\phi^{ens \rightarrow m }{i, j} = f{e \in E_i} (norm(\phi^{e\rightarrow m}_{i, j})).
\end{equation}

An important fact needs to be emphasized here. The first two normalization approaches center the explanation data, which may change the attribution's meaning. Signs of the feature's attribution might be changed after the normalization.

\begin{figure}[h]
  \centering
  \includegraphics[width=0.6\linewidth]{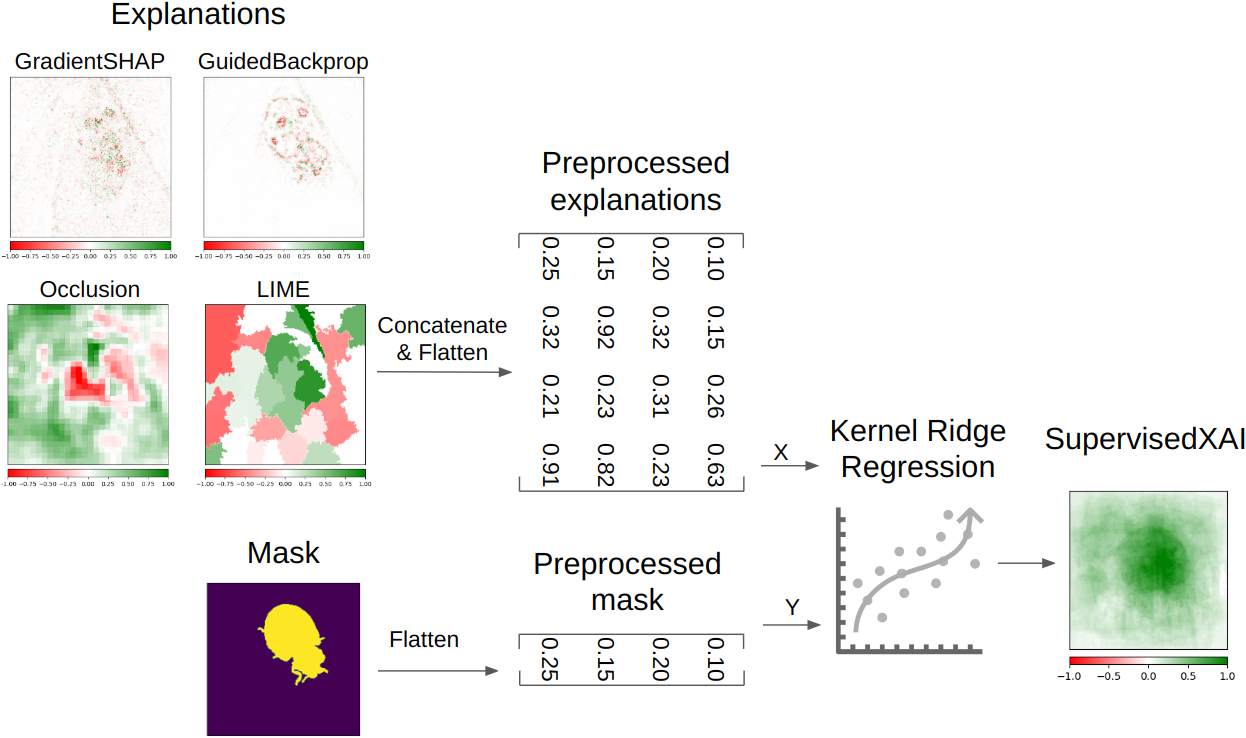}
  \caption{Visualization of the SupervisedXAI method~\cite{Zou2022}. Explanations for an instance are reshaped and concatenated into a matrix, which constitutes the training dataset X. The segmentation mask is similarly reshaped into a one-dimensional vector, serving as a set of labels Y. A multioutput Kernel Ridge Regression (KRR) model is then trained to predict the values of Y, using the explanations transformed into the X matrix as input.}
\end{figure}

\section{State-of-the-art XAI ensembling methods}

\paragraph{SupervisedXAI} is an algorithm proposed by~\citet{Zou2022}, which requires ground truth segmentation masks. The explanations of an instance are reshaped and concatenated to a~one-dimensional vector. The segmentation mask is also reshaped to a~one-dimensional vector. These vectors are concatenated vertically to form matrix $X$ and matrix $Y$ accordingly. A multi-output kernel ridge regression (KRR) is trained to predict $Y$ values based on explanations transformed into $X$ matrix. Either k-fold cross-validation is used to prevent data leakage, or the data is split into training and testing subsets. 

We propose an improvement of the algorithm. We use instance weights in fitting the KRR to differentiate the impact of big and small masks in fitting the model. A heuristic is used to determine the weight; it is inversely proportional to the~area of the segmentation mask. All input explanations are scaled by the~average second-moment estimate before fitting the~KRR model to increase numerical stability.

\paragraph{Autoweighted} is an explanations ensembling method, in which explanations are evaluated using chosen metrics, and then each explanation's Ensemble Score (ES)~\cite{Bobek2021} is calculated. The final ensemble is performed as a~weighted mean of normalized explanations with individual weights equal to the~ES.

In this work, stability and consistency are the~metrics used to calculate the~ES. Stability measures how the~explanation changes when the~input image is perturbed with random Gaussian noise. Consistency measures the~change in explanation when the~machine learning model is changed. 

\begin{figure}[h]
  \centering
  \includegraphics[width=0.6\linewidth]{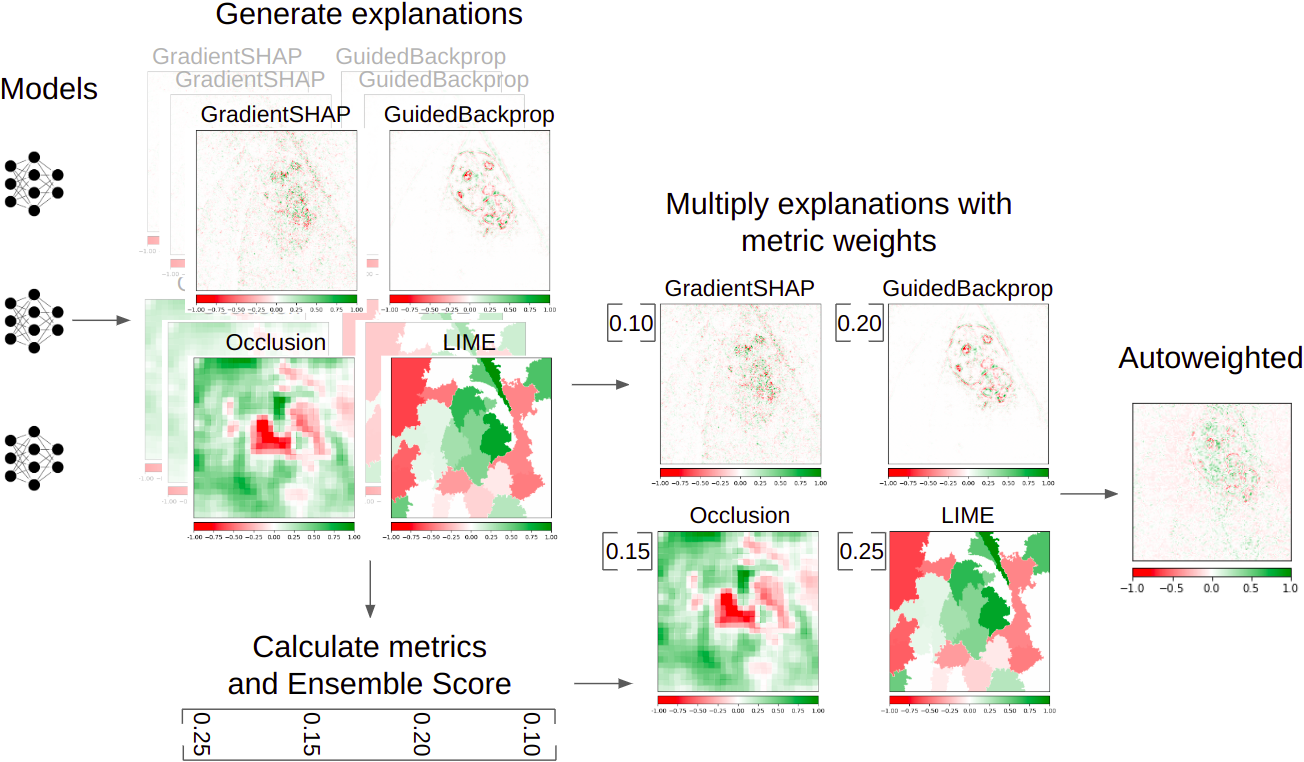}
  \caption{Illustration of the Autoweighted method~\cite{Bobek2021}. In this approach, explanations are assessed using a chosen metric, and subsequently, an Ensemble Score (ES) is computed for each explanation method. The final XAI ensemble is constructed as a weighted mean of normalized explanations, with individual weights determined by their respective ES values.}
\end{figure}

\section{Experiments}

\textbf{Experimental setup} Experiments were conducted using the~ImageNet-S dataset~\cite{gao2022luss}. The subset of 50 classes of 10 instances with segmentation masks was used. The data was divided into training and testing with a~4:1 ratio and in-class stratification. The pretrained on ImageNet~\cite{5206848} Resnet-18 model was used to generate the~compositional explanations. Using the~Captum library~\cite{Kokhlikyan2020}, twelve explanations were generated, such as Integrated Gradients, Saliency, Gradient SHAP, Guided Backpropagation, Deconvolution, Input X Gradient, LIME, Occlusion, Shapley Value Sampling, Feature Ablation, KernelSHAP, Noise Tunnel. The quality of the~ensemble explanation was assessed in 5 categories~\cite{Hedstrom2022}: Faithfulness, Robustness, Localization, Complexity, Randomization by the~most relevant metric by~\citet{hedstrom2023metaquantus}. 

\begin{table*}[h]
\centering
\begin{tabular}{@{}p{2cm}p{2cm}lllll@{}}
\toprule
Normalization & \shortstack{Aggregating\\function} & \textbf{Fa} $\downarrow$ & \textbf{Ra} $\downarrow$ & \textbf{Ro} $\downarrow$ & \textbf{Co} $\uparrow$ & \textbf{Lo} $\uparrow$ \\ \midrule
\multirow{4}{2cm}{Normal standardization} & max & \textbf{0.022} & 0.590 & 0.207 & 0.359 & 0.670 \\
& min & 0.058 & 0.631 & 0.184 & 0.304 & 0.638 \\
& \textbf{avg} & 0.023 & \textbf{0.107} & 0.325 & 0.462 & \textbf{0.790} \\
& max\_abs & 0.024 & 0.682 & 0.196 & 0.275 & 0.700 \\ \cline{2-7}
\multirow{4}{2cm}{Robust standardization} & max & 0.064 & 0.480 & 0.862 & 0.778 & 0.750 \\
& min & 0.072 & 0.503 & 0.816 & \textbf{0.823} & 0.635 \\
& avg & 0.061 & 0.344 & 1.064 & 0.728 & 0.700 \\
& max\_abs & 0.066 & 0.374 & 0.956 & 0.692 & 0.710 \\ \cline{2-7}
\multirow{4}{2cm}{Second Moment Scaling} & max & \textbf{0.022} & 0.582 & 0.218 & 0.311 & 0.660 \\
& min & 0.078 & 0.599 & 0.189 & 0.364 & 0.570 \\
& \textbf{avg} & 0.023 & 0.167 & 0.314 & 0.469 & \textbf{0.790} \\
& max\_abs & 0.023 & 0.658 & \textbf{0.201} & 0.276 & 0.680 \\ \hline
\end{tabular}
\caption{The quality of NormEnsembleXAI ensemble evaluated on ImageNet and COCO dataset. Comparison of different aggregating functions and normalization methods. We evaluated the ensemble explanation across different metrics: Faithfulness (Fa) is assessed using Pixel-Flipping, Randomization (Ra) through Random Logit, Robustness (Ro) via Local Lipschitz Estimation, Complexity (Co) using Sparseness, and Localization (Lo) determined with the Pointing-Game method.}
\label{tab:normalization_comparison}
\end{table*}

\subsection{Ablation studies}

The motivation behind conducting ablation studies lies in the necessity to unravel the intricacies of XAI ensemble methods, particularly when it comes to the selection of hyperparameters. The freedom to choose hyperparameters introduces a realm of possibilities, and it becomes imperative to discern the impact of various normalization methods and aggregation functions on the overall performance of the NormEnsembleXAI.

Ablation studies, as presented in Table~\ref{tab:normalization_comparison}, systematically explore different combinations of normalization algorithms and aggregation functions. This rigorous exploration allows us to gauge the effectiveness of each approach across diverse evaluation metrics. 

Identifying one best method depends strongly on the~metric choice. For instance, when prioritizing Complexity (Sparseness), robust standardization emerges as the top-performing normalization method. On the~other hand, the~Local Liepschitz Estimate shows that median-IQR normalization, contrary to the~name, is the~least robust in the~XAI ensemble setting. We attribute this result to the~fact that median and IQR are not smooth functions, and their Local Lipschitz Estimate is high. Approaches aggregating by the~average value with second-moment scaling and normal standardization strike a~balance in all metrics. 

In conclusion, approaches employing average aggregation with second-moment scaling and normal standardization strike a balance across all metrics, showcasing their versatility and effectiveness across diverse scenarios. This nuanced understanding of how normalization methods interact with aggregation functions could guide practitioners in making rational decisions when configuring NormEnsembleXAI for specific use cases.

\subsection{Comparative analysis}

In our quest to unveil the strengths and weaknesses of XAI ensemble methods, we conducted a comprehensive evaluation, as depicted in Table~\ref{tab:ensemble_comparison}. The table presents a comparative assessment of the~quality of XAI ensembles on both ImageNet and COCO datasets, encompassing key evaluation metrics.

\begin{table*}
\centering
\begin{tabular}{@{}p{5.0cm}llllll@{}}
\toprule
& Fa $\downarrow$  & Ra $\downarrow$ & Ro $\downarrow$  & Co $\uparrow$ & Lo $\uparrow$ \\ \midrule
Component explanations & 0.051 & 0.367 & 0.549 & \textbf{0.553} & 0.720 \\
Autoweighted & 0.024 & 0.344 & 0.249 & 0.526 & 0.760 \\
SupervisedXAI no weights & 0.104 & 0.757 & \textbf{0.217} & 0.393 & 0.760 \\ \midrule
SupervisedXAI auto weights & 0.102 & 0.735 & 0.227 & 0.395 & \textbf{0.810} \\
\textbf{NormEnsembleXAI normal avg} & \textbf{0.023} & \textbf{0.107} & 0.325 & 0.462 & 0.790 \\
\textbf{NormEnsembleXAI scaling avg} & \textbf{0.023} & 0.167 & 0.314 & 0.469 & 0.790 \\ \bottomrule
\end{tabular}
\caption{Assessment of ensemble explanation quality on ImageNet and COCO datasets. The presented numbers represent the mean values of evaluation metrics across the entire dataset, with errors calculated as the standard deviation of metric scores. For reference, we include the mean (across the~dataset) values of component explanations. We evaluated ensemble explanation across different metrics: Faithfulness (Fa) is assessed using Pixel-Flipping, Randomization (Ra) through Random Logit, Robustness (Ro) via Local Lipschitz Estimation, Complexity (Co) using Sparseness, and Localization (Lo) determined with the Pointing-Game method.
}
\label{tab:ensemble_comparison}
\end{table*}

For clarity, only the~two most promising ensemble methods from the~NormEnsembleXAI group are shown: aggregation by average with the~average second moment estimate normalization and aggregation by average with normal standardization. Both variants, with uniform weights and automatic weights, of the~SupervisedXAI were evaluated. Since their results are nearly identical, only the~results of the~version with automatic weights are presented.

The results of the~Autoweighted method are comparable to the~results of NormEnsembleXAI algorithms. The slightly better complexity and robustness of Autoweighted can be traded for a~more desirable score in the~randomization category. The similarity of these methods can explain this proximity of scores. Autoweighted uses a~weighted average to ensemble explanations. When the~Ensemble Scores of component explanations are similar, the~method acts like the~NormEnsembleXAI with aggregation by average.

\begin{figure}[h]
    \centering
    \includegraphics[width=0.8\linewidth]{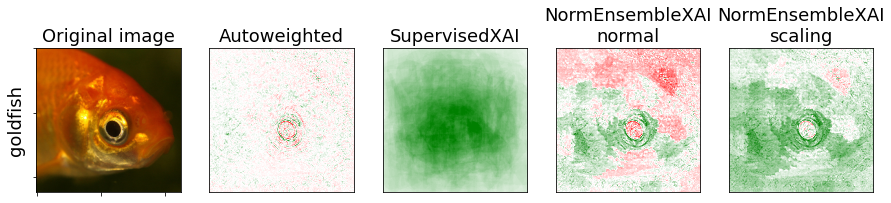}
    \includegraphics[width=0.8\linewidth]{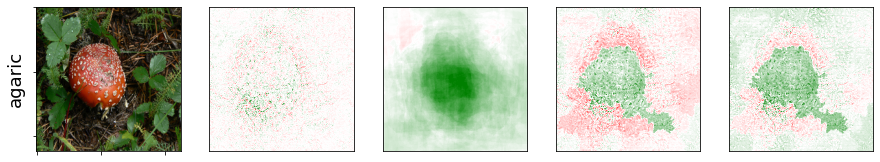}
    \includegraphics[width=0.8\linewidth]{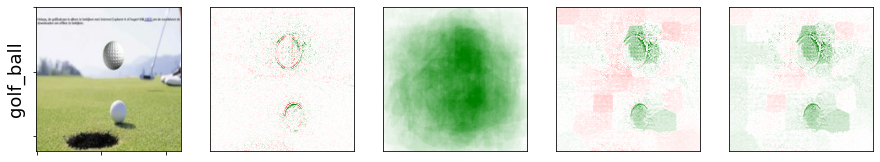}
    \includegraphics[width=0.8\linewidth]{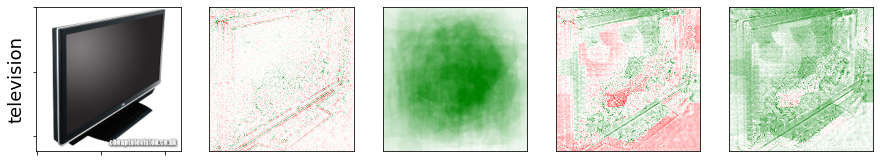}
    \includegraphics[width=0.8\linewidth]{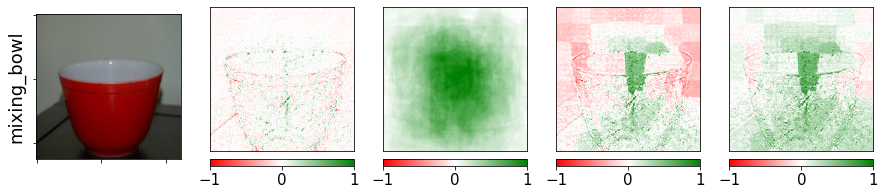}
    \caption{Examples of XAI ensembling results. Each column displays the following images: the original image, XAI ensembling obtained with the supervised method, XAI ensembling obtained with the Autoweighted method, and our result using NormEnsembleXAI. In our approach, we employ average aggregation and utilize two distinct normalization techniques: Normal Standardization and Second Moment Scaling.}
    \label{fig:ensembling_examples}
\end{figure}

Localization performance measured by the~Pointing-Game metric is statistically the~same for all ensemble methods. It means that, on average, the~pixel with the~highest attribution is within the~segmentation mask the~same number of times for each ensemble method.

In every category of metrics exists a~component explanation algorithm that yields better results than the~ensemble methods. However, the~best method changes from metric to metric. The main benefit of explanations ensembling is their universality and satisfactory performance in every category. 

The only metric where component explanations score visibly better than the~ensembles is Sparseness in the~Complexity category. This result is unsurprising: the~ensemble explanation depends on many single explanations and tends to be less sparse.
    
In conclusion, our analysis provides a comprehensive comparison of the XAI ensemble methods, shedding light on their versatility and consistent performance across multiple evaluation metrics. The effectiveness of these methods is exemplified in Figure~\ref{fig:ensembling_examples}, which shows visual examples of XAI ensembling results on diverse inputs from the ImageNet dataset.

\subsection{Limitations of ensembling XAI methods}

We conducted experiments that allowed us to draw conclusions about the weakness of XAI ensembling methods. These limitations have surfaced during our experimentation and analysis, revealing challenges and constraints in harnessing the~full potential of XAI ensembling techniques.

\paragraph{Time consumption}
We decided to look at the cost-effectiveness of the time aggregation method. Compared to other methods, presented in Table~\ref{tab:time}, Autoweighted performs the worst. This is due to the need to calculate Ensemble Score. 

Next in line is SupervisedXAI, the complexity of which is strongly correlated with the number of images for which explanations were generated and then the KRR model was trained on them. Usage of the~SupervisedXAI method on a~large scale is also limited by the~memory complexity of the~algorithm. All explanations and masks must first be loaded into memory, and only then a~KRR model can be fitted. To our best knowledge, there does not exist a~batched version of the~KRR algorithm which could potentially solve this issue.

The NormEnsembleXAI method performs best in terms of time, for which the explanation generation time on the NVIDIA A100-SXM4-40GB card is less than 0.12 seconds.

\begin{table}
\centering
\begin{tabular}{l c}
\toprule
\textbf{Method} & \textbf{Average time (s)} \\
\midrule
Autoweighted & $48.699 \pm 3.297$ \\
SupervisedXAI(500 samples) & $41.858 \pm 0.028$ \\
SupervisedXAI(20 samples) & $0.972 \pm 0.046$ \\
NormEnsembleXAI normal avg  & $0.114 \pm 0.078$ \\
NormEnsembleXAI scaling avg  & $0.088 \pm 0.059$ \\
\bottomrule
\end{tabular}
\label{tab:time}
\caption{Average time (in seconds) of ensembling explanations without generating component explanations. The average time of the SupervisedXAI method depends on the number of samples while training the KRR model. For more samples, the time will be longer.}
\end{table}

\paragraph{Possibility of bias}

During generation of the~SupervisedXAI, we discovered that the~ensemble explanations have some very distinctive properties. Namely, the~area with the~highest attribution was in the~center of the~image, and the~attribution was close to 0 near the~edges. We assume that decreasing the~number of component explanations and, consequently, the~number of features passed to the~KRR model may increase the~model's bias and lead to even worse ensemble performance.

In case of Autoweighted, we followed the~original paper authors' guidelines regarding the~choice of metrics used to calculate the~Ensemble Score. However, it does not always need to be the~best approach. The responsibility for the~proper metric selection lies with the~researcher/analyst, which may introduce bias towards selected metric.

Compared to other methods, the possibility of bias of NormEnsembleXAI is relatively minor. The results of this algorithm are significantly influenced by the choice of the aggregating function. Although our experiments have consistently demonstrated that aggregation by mean yields the highest scores, we acknowledge that there may be situations where averaging could be potentially misleading. A noteworthy example arises when two component explanations provide attributions with equal magnitude but opposite signs, leading to undesirable information loss through cancellation.

\paragraph{Requirement of additional resources}

The necessity of pixel-level annotations poses a significant hurdle for the SupervisedXAI method, as such annotations are uncommon, and even when available, only a few instances are typically annotated. Autoweighted requires multiple models, posing a challenge when working with publicly available models not authored by the user. In contrast, NormEnsembleXAI requires no additional resources, other than for component explanations.

\paragraph{Only positive feature attributions}
We identify another, yet minor, limitation of the~SupervisedXAI method. Due to the~fact that annotation masks are usually single-channel, the~ensemble explanation also has one channel. The values of the~result of SupervisedXAI are also limited by the~values present in the~mask. Usually, masks are binary, so the~ensemble explanation is bounded to the~$[0,1]$ interval. It limits the~possibility of expressing negative attributions. Although it is not our recommended configuration, the~NormEnsembleXAI can also produce positive values alone with maximum aggregation or negative values alone with minimum aggregation. The problem of negative contributions does not occur with the~Autoweighted. \newline

In conclusion, our experiments on XAI ensembling methods have brought to light several critical limitations and considerations. Time consumption analysis revealed Autoweighted's significant computational demands, while SupervisedXAI faced challenges related to bias and attribution limited to positive values. NormEnsembleXAI emerged as a standout performer with minor constrain related to the sensitivity of selection of aggregating function.

It is worth noting that NormEnsembleXAI demonstrated exceptional performance in three out of four tested limitations. It exhibited superior time efficiency, minimal bias concerns compared to other methods, and a resource-friendly approach without the need for any additional resources beyond component explanations. These results position NormEnsembleXAI as a reliable and versatile choice in the realm of XAI ensembling techniques, offering a promising solution for researchers and practitioners seeking effective and efficient model interpretability.

\section{XAI ensembling Python library}
Explanations ensembling algorithms introduced in Section~\ref{sec:XAI_ensemblings} were implemented in a~new, open-source, Python library EnsembleXAI. It is PyTorch~\cite{NEURIPS2019_9015} and Captum~\cite{Kokhlikyan2020} compatible and has extensive documentation. The library consists of three modules: Ensemble, which provides explanation ensembling functionalities; Metrics, which contains various explainability metrics, primarily those introduced in~\cite{Zou2022, Bobek2021}; and Normalization, with scaling and standardizing functions. The library is available at \href{https://github.com/Hryniewska/EnsembleXAI}{https://github.com/Hryniewska/Ensemble\_XAI}.

\section{Conclusions and Future Work}

The scores achieved during Quantus evaluation show the~benefits of using an ensemble in an explanatory analysis of DL models. Conducted experiments present that the~straightforward approach to ensemble explanations can give comparable results to more complicated ensembling methods. An important part of successfully combining multiple explanations is normalization. 

Limitations of ensemble algorithms pose challenges in the~field of XAI, which may be resolved in the~future. We think that the~SupervisedXAI method can be improved by using convolutional neural networks (CNNs) instead of the~KRR model. CNNs are a~good tool for extracting spatial relationships in the~images, and we hope it could be leveraged to construct a~better ensembling algorithm.

Furthermore, there is a need to explore the generalizability of our proposed method to various data types, including text and tabular data. Extending our methodology to these domains will be a relevant area of future investigation.

\section{Acknowledgments}

We would like to thank Mateusz Sperkowski for his contribution to the~preparation of the~EnsembleXAI library. 
This work was carried out with the~support of the~Laboratory of Bioinformatics and Computational Genomics and the~High Performance Computing Center of~the~Faculty of Mathematics and Information Science Warsaw University of Technology.

\bibliographystyle{IEEEtranN}
\bibliography{main}

\end{document}